\pdfoutput=1

\documentclass[11pt]{article}

\usepackage[]{EMNLP2023}

\usepackage{times}
\usepackage{latexsym}

\usepackage[T1]{fontenc}

\usepackage[utf8]{inputenc}

\usepackage{microtype}

\usepackage{inconsolata}

\usepackage{multirow}
\usepackage{booktabs}
\usepackage{lipsum}
\usepackage{graphicx}
\usepackage{amsmath}
\usepackage{amssymb}
\usepackage{enumitem}
\usepackage{lipsum}
\usepackage{makecell}

\setlist{topsep=3pt,itemsep=3pt,partopsep=3pt, parsep=3pt}

\title{InstructTODS: Large Language Models for End-to-End Task-Oriented Dialogue Systems}

\author{
  Willy Chung, Samuel Cahyawijaya, Bryan Wilie, Holy Lovenia, Pascale Fung \\
  Hong Kong University of Science and Technology \\
  Clear Water Bay, Hong Kong \\
  \texttt{whcchung@connect.ust.hk} \\ 
}

\begin{document}
\maketitle
\begin{abstract}


Large language models (LLMs) have been used for diverse tasks in natural language processing (NLP), yet remain under-explored for task-oriented dialogue systems (TODS), especially for end-to-end TODS. We present InstructTODS, a novel off-the-shelf framework for zero-shot end-to-end task-oriented dialogue systems that can adapt to diverse domains without fine-tuning. By leveraging LLMs, InstructTODS generates a proxy belief state that seamlessly translates user intentions into dynamic queries for efficient interaction with any KB. Our extensive experiments demonstrate that InstructTODS achieves comparable performance to fully fine-tuned TODS in guiding dialogues to successful completion without prior knowledge or task-specific data. Furthermore, a rigorous human evaluation of end-to-end TODS shows that InstructTODS produces dialogue responses that notably outperform both the gold responses and the state-of-the-art TODS in terms of helpfulness, informativeness, and humanness. Moreover, the effectiveness of LLMs in TODS is further supported by our comprehensive evaluations on TODS subtasks: dialogue state tracking, intent classification, and response generation. Code and implementations could be found here\footnote{\url{https://github.com/WillyHC22/InstructTODS/}}.
\end{abstract}

\section{Introduction}



LLMs have consistently pushed new frontiers in natural language processing (NLP) in terms of performance across a variety of benchmarks, such as MMLU \cite{hendrycksmeasuring}, BIG-Bench \cite{lewkowycz2022beyond} and HELM \cite{bommasani2023holistic}, achieving state-of-the-art results in both natural language understanding (NLU) and generation (NLG) tasks \cite{bang2023multitask}. 
Various applications of LLMs have also been adopted in the industry, most prominently ChatGPT\footnote{\url{http://chatgpt.openai.com/}} and GPT-4\footnote{\url{https://openai.com/gpt-4}}, which can provide a natural answer to a diverse range of questions fluently and coherently.

\begin{figure}[!t]
  \centering
  \resizebox{0.95\linewidth}{!}{
  \includegraphics{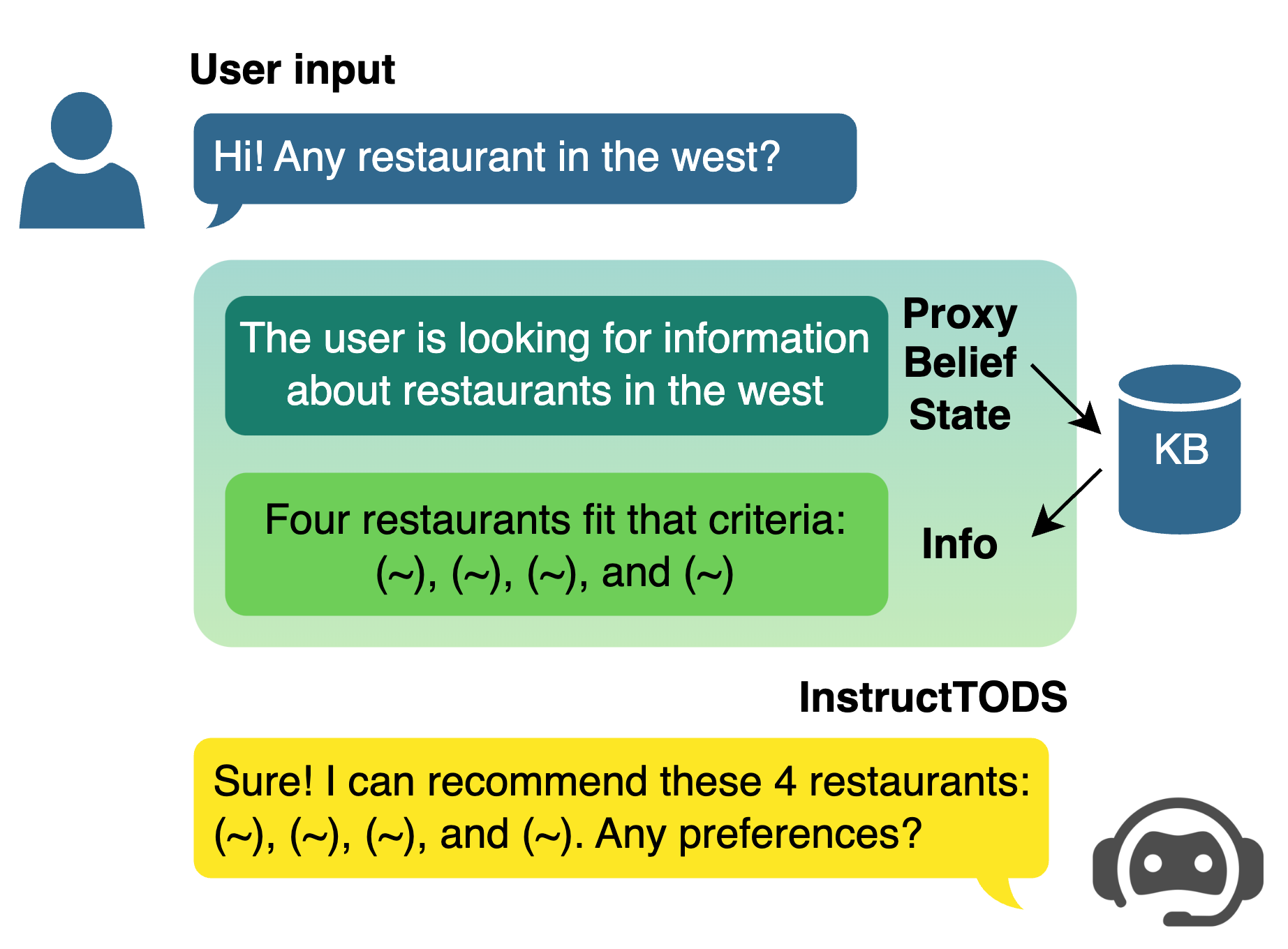}
  }
  \caption{InstructTODS is the first zero-shot end-to-end task-oriented dialogue system that requires no task-specific annotations, and ontology while generating more human-preferred responses.}
  \label{fig:figure1}
\end{figure}

\begin{table*}[!t]
\centering
\resizebox{0.8\linewidth}{!}{
    \begin{tabular}{l|l|c|ccc|c|ccc|c}
    \toprule
    \multicolumn{1}{c|}{\multirow{3}{*}{\textbf{Setting}}} & \multirow{3}{*}{\textbf{Model}} & \multicolumn{4}{c|}{\textbf{Banking77}} & \multicolumn{5}{c}{\textbf{CLINC150}}\\ 
    \cmidrule{3-6} \cmidrule{7-11}
    & & \multirow{2}{*}{\textbf{Single}} & \multicolumn{3}{c|}{\textbf{Multi}} & \multirow{2}{*}{\textbf{Single}} & \multicolumn{3}{c|}{\textbf{Multi}} & \multirow{2}{*}{\textbf{OOS}*} \\ 
    & & & top-1 & top-2 & top-3 & & top-1 & top-2 & top-3 & \\
    \midrule
    & RoBERTa$_{large}$ & 78.99 & -- & -- & -- & 89.89 & -- & -- & -- & -- \\
    \textbf{Few-Shot} & ICDA$_{XL}$ & 83.90 & -- & -- & -- & 92.62  & -- & -- & -- & -- \\
    \textbf{Baseline} & DNNC & 80.40 & -- & -- & -- & 91.02  & -- & -- & -- & -- \\
    & CPFT & 80.86 & -- & -- & -- & 92.34  & -- & -- & -- & -- \\
    \midrule
    \textbf{Zero-Shot} & BART$_{large}$-MNLI & 35.91 & 35.91 & 49.09 & 56.14 & 26.44 & 26.44 & 36.46 & 43.13 & 0.2\\
    \textbf{Baseline} & BART$_{large}$-MNLI$_{corr}$ & 42.24 & 42.24 & 55.19 & 62.2 & 40.16 & 40.16 & 51.88 & 57.66 & 0.9\\
    \midrule
    \textbf{Zero-Shot} & Modular (GPT-3.5) & 65.45 & \textbf{64.51} & 77.69 & 83.18 & 64.91 & 63.22 & 72.20 & 82.42 & 10.9 \\
    \textbf{LLM} & Modular (GPT-4) & \textbf{74.09} & 64.06 & \textbf{80.75} & \textbf{86.33} & \textbf{73.91} & \textbf{69.90} & \textbf{81.88} & \textbf{90.33} & \textbf{62.1} \\
    \bottomrule
    \end{tabular}
}
\caption{Performance comparison on \textbf{intent classification}. LLMs outperform most baselines in our benchmark. Best performances in each section are in \textbf{bold}. *Out-of-scope intent of CLINC150.}
\label{tab:ic_results}
\end{table*}


Among the manifold tasks in NLP, task-oriented dialogue systems (TODS) represent a crucial domain.
In general, TODS can be categorized into: the pipelined approach \cite{ham2020end,hosseini2020simple,ohashi2022post}, relying on multiple sequential modules and heavy annotations for dialogue states and system actions, and the end-to-end approach \cite{banerjee2019graph,qin2020dynamic,he2022galaxy}, where the systems generate responses directly from the user input and the KB. Both approaches lack adaptability to unseen domains. This adaptability often requires domain-specific structures (ontology), and data for TODS is notoriously expensive to collect and annotate \cite{eric2020multiwoz}. In this regard, LLMs present great potential thanks to their extensive pre-trained knowledge, enabling them to adapt to contextual information without any parameter updates or additional task-specific data.

However, utilizing LLMs for tasks requiring knowledge grounding, such as 
TODS, poses a critical challenge that calls for thorough investigation and exploration. TODS requires dialogue systems to adeptly complete a specific goal by interacting with a user in natural language according to a certain set of bounded functions, ontology, and knowledge within the corresponding domain. Nevertheless, naively feeding all of the knowledge to the LLMs in TODS could lead to the generation of misleading and unfaithful information, i.e., hallucination~\cite{ji2023survey,azamfirei2023large}.

In this work, we first investigate the capability of LLMs to perform three key TODS objectives in zero-shot settings, specifically dialogue state tracking (DST), intent classification (IC), and response generation (RG). While LLMs demonstrate impressive capabilities and understanding of these tasks individually, a closer examination of their shortcomings reveals that the modular approach is not the most suitable for effectively using LLMs in TODS due to its restrictiveness. Rather than confining interactions within predefined elements like slots, values, or system actions, it is more advantageous to harness the emergent abilities of LLMs to process unstructured information, which also enables the system to easily adapt to new domains.

From these observations, we propose InstructTODS, a fully off-the-shelf framework to perform end-to-end unified TODS in a zero-shot setting using LLMs. InstructTODS is adaptable to any KB and does not require any ontologies or task-specific data. Instead of using predefined slot values, InstructTODS generates an unstructured proxy belief state from the dialogue context. Then, an action thought is generated to query the KB dynamically in natural language using an LLM. The retrieved information is then given to generate the response.



%


In summary, our contributions are as follows:
\begin{itemize}
    \item We provide an extensive evaluation and comprehensive analysis of LLMs’ zero-shot performance in several TODS subtasks, notably intent classification, dialogue state tracking, and response generation.
    
    \item We introduce InstructTODS, a fully off-the-shelf framework to leverage instruction-tuned LLMs in zero-shot setting for end-to-end unified task-oriented dialogue, with the benefit of being effectively adaptable to any knowledge base (KB) while alleviating the need for any additional form of task-relevant data, such as intent, belief state, system action, etc.
    
    \item We provide valuable insights from the TODS experiments on the more general advantages and failure cases of LLMs to perform complex zero-shot NLP tasks.
    
\end{itemize}

\begin{table*}[!t]
\centering
\resizebox{1.0\linewidth}{!}{
\begin{tabular}{l|cc|cc|cc|cc|cc|cc}
\toprule
\multirow{2}{*}{\textbf{Model}} & \multicolumn{2}{c|}{Attraction} & \multicolumn{2}{c|}{Hotel} & \multicolumn{2}{c|}{Restaurant} & \multicolumn{2}{c|}{Taxi} & \multicolumn{2}{c|}{Train} & \multicolumn{2}{c}{\textbf{Average}} \\
& \textbf{JGA} & \textbf{Slot-F1} & \textbf{JGA} & \textbf{Slot-F1} & \textbf{JGA} & \textbf{Slot-F1} & \textbf{JGA} & \textbf{Slot-F1} & \textbf{JGA} & \textbf{Slot-F1} & \textbf{JGA} & \textbf{Slot-F1} \\
\midrule
TRADE & 20.06 & -- & 14.20 & -- & 12.59 & -- & 59.21 & -- & 22.39 & -- & 25.69 & -- \\ 
MA-DST & 22.46 & -- & 16.28 & -- & 13.56 & -- & 59.27 & -- & 22.76 & -- & 26.87 & -- \\ 
TransferQA & 31.25 & -- & 22.72 & -- & 26.28 & -- & 61.87 & -- & 36.72 & -- & 35.77 & --  \\
T5Dep & 37.83 & -- & 26.50 & -- & 27.05 & -- & \textbf{69.23} & -- & 40.27 & -- & 40.18 & -- \\ 
\midrule
Modular (GPT-3.5, w/ all slots) & 30.23 & 65.38 & 26.77 & 76.28 & 48.28 & 82.90 & 56.22 & 75.33 & 53.75 & 83.64 & 42.02 & 78.60 \\
Modular (GPT-3.5, w/ domain slot) & 39.53 & 74.89 & 27.03 & 79.78 & 51.72 & 85.06 & 63.24 & 83.98 & 52.50 & 84.84 & 44.48 & 82.53 \\
Modular (GPT-4, w/ all slots) & 39.53 & 78.99 & 31.23 & \textbf{84.07} & 55.86 & 88.23 & 63.24 & 82.71 & \textbf{59.83} & 89.72 & 48.16 & 85.62 \\
Modular (GPT-4, w/ domain slot) & \textbf{46.51} & \textbf{81.13} & \textbf{31.76} & 83.42 & \textbf{56.90} & \textbf{88.47} & \textbf{65.96} & \textbf{84.33} & 52.50 & \textbf{89.73 }& \textbf{48.35} & \textbf{85.82} \\
\bottomrule
\end{tabular}
}
\caption{Performance comparison on \textbf{zero-shot DST benchmark}. LLMs outperform all baselines in our benchmark. Baseline results are directly taken from their respective works. The best performances in each section are in \textbf{bold}.}
\label{tab:dst_result}
\end{table*}

\section{Evaluating LLMs on Zero-Shot Task-Oriented Dialogue Subtasks}

As an intermediary step in exploring the potential of end-to-end TODS solutions, we first investigate how well the performance of state-of-the-art LLMs (we presented the comparison of different LLMs over multiple tasks in Appendix~\ref{app:llm-comparison}), i.e., GPT-3.5 and GPT-4, in performing various modular task-oriented objectives in their respective settings.

\subsection{TODS Subtasks} 

Let us define a dialogue set $\mathcal{D}_n = \{u_1, r_1, u_2, r_2, ..., u_n, r_n\}$ where $u_i$ and $r_i$ denotes the user utterance and the system reply at turn $i$, respectively.

\paragraph{Intent Classification (IC)}
For IC, we have the set of labels $ C = \{c_1, c_2, ..., c_t\}$, from which we build the input for the LLM as $x^{ic}_{i} = \mathbb{P}^{ic}(\mathcal{I}^{ic}, Concat(c_j)_{j=0}^{t}, u_i)$ where $\mathbb{P}^{ic}(.)$ is the IC input template, $\mathcal{I}^{ic}$ refers to the natural language instruction for IC and $Concat(c_j)$ is the concatenation of all labels. We evaluate two generation settings, a single output setting where we query the model for the inferred intent, and a multi-output setting where we query the model for the top-3 intents given the user query by simply changing the instruction $\mathcal{I}^{ic}$. As such, we recast the classification task in a text-generation manner and compare our results with state-of-the-art IC baselines. 

\paragraph{Dialogue State Tracking (DST)}
For DST, we define the total set of slots $S = \{s_{1,D_1}, s_{2, D_1}, ..., s_{k, D_l}\}$ where $s_{i, D_j}$ is the i-th slot associated to domain $D_j$. We give a singular hand-crafted exemplar distinct from the dataset to guide the generation format directly as JSON. We build the input $x^{dst}_i = \mathbb{P}^{dst}(\mathcal{I}^{dst}, f^{dst}(S), \mathcal{D}_i)$ by providing the entire dialogue context, where $\mathbb{P}^{dst}(.)$ is the DST input template, $\mathcal{I}^{dst}$ denotes the instruction for DST $f^{dst}(S)$ refers to a textual transformation of the set of slots. We evaluate two settings with different slot transformations: one by providing all slots and another with only the active domain slots.

\paragraph{Response Generation (RG)}
For RG, given a dialogue $\mathcal{D}$, we define the set of oracle system actions $A = \{a_{1,1}, a_{1,2}, ..., a_{n, m}\}$ where $a_{i, j}$ denotes the j-th system action of turn i. We construct the input $x^{rg}_i = \mathbb{P}^{rg}(\mathcal{I}^{rg}, f^{rg}(a_{i, 1}, a_{i, 2}, ..., a_{i, m}), \mathcal{D}_i)$ where $\mathbb{P}^{rg}(.)$ is RG input template, $\mathcal{I}^{rg}$ denotes the instruction for RG and $f^{rg}(.)$ refers to a textual transformation of the set of system actions. We evaluate the capability of LLMs to leverage a structured system action while addressing the dialogue context to generate a response to the user.


\subsection{Experiment Settings}

\paragraph{Dataset} For the dialogue state tracking, we evaluate the LLMs' capability on MultiWOZ 2.1 (MWOZ) \cite{eric2020multiwoz}. For intent classification, we evaluate two datasets: Banking77 \cite{casanueva2020efficient}, a fine-grained intent dataset in the banking domain, and CLINC150 \cite{larson2019evaluation}, coarse-grained intents classification datasets covering over 10 different domains. The main challenge of the CLINC150 dataset is on inferring out-of-scope intent, which is particularly challenging without any model training.

\paragraph{Evaluation} We evaluate dialogue state tracking with Joint Goal Accuracy (JGA) and Slot-F1. We compute JGA using exact matching instead of fuzzy matching, with minor typo fixes in MWOZ following prior works~\cite{hosseini2020simple,su2022multi}. For intent classification, we evaluate the accuracy when predicting only one intent (single) and the top-3 intents (multi) in a text generation setting. BLEU~\cite{papineni2002bleu}, Inform, and Success \cite{eric2020multiwoz} are used for response generation. In addition to these metrics, we also compare lexical diversity~\cite{shen2022lexicalrichness}, i.e., HDD~\cite{mccarthy2010mtldhdd}, MATTR~\cite{covington2010mattr}, MTLD~\cite{mccarthy2005mtld}, and VOCD~\cite{mccarthy2007vocd}, fluency through perplexity, and human-likability using USL-H~\cite{phy2020uslh}.

\begin{table*}[!t]
\centering
\resizebox{0.85\linewidth}{!}{
\begin{tabular}{l|ccc|c|cccc|c}
\toprule
\multicolumn{1}{c}{\multirow{2}{*}{\textbf{Model}}} & \multicolumn{3}{|c}{Reference-based} & \multicolumn{1}{|c}{Fluency} & \multicolumn{4}{|c}{Lexical diversity} & \multicolumn{1}{|c}{Human pref.} \\
& \textbf{BLEU} & \textbf{Inf.} & \textbf{Succ.} & \textbf{PPL} & \textbf{HDD} & \textbf{MATTR} & \textbf{MTLD} & \textbf{VOCD} & \textbf{USL-H} \\
\midrule
LAVA    & 11.33 & 95.8 & \textbf{94.9 }& \textbf{25.45} & 65.35 & 74.84 & 30.72 & 25.84 & 59.68 \\
SFN     & 14.11 & \textbf{97.7} & 91.6 & 51.97 & 70.68 & 78.67 & 34.25 & 36.02 & 65.41 \\
DAMD    & 14.94 & 78 & 68.7 & 58.41 & 71.45 & 78.09 & 29.08 & 37.57 & 62.62 \\
MARCO   & 16.5 & 95.3 & 91.1 & 36.00 & 73.40 & 83.39 & 44.48 & 42.78 & 70.35 \\
MinTL   & 18.39 & 85 & 80.8 & 49.77 & 71.31 & 82.76 & 38.99 & 37.26 & 65.36 \\
BORT    & 16.75 & 91.1 & 88.3 & 53.45 & 70.94 & 81.82 & 38.41 & 36.28 & 66.00 \\
HDSA    & \textbf{20.02} & 95.8 & 90.2 & 43.37 & 71.71 & 82.95 & 42.04 & 38.02 & 68.36 \\
RSTOD   & 15.98 & 91.6 & 86.9 & 76.05 & 73.11 & 82.41 & 42.08 & 41.88 & 68.54 \\
\midrule
Modular (GPT-4) & 6.12 & 86.42 & 78.48 & 36.63 & \textbf{80.59} & \textbf{89.56} & \textbf{66.64} & \textbf{70.13} & \textbf{89.66} \\
\bottomrule
\end{tabular}
}
\caption{Performance comparison on \textbf{response generation}. Although lower in BLEU, responses by the LLM-powered modular TODS are more human-preferred. The reported results for the baselines are taken from their respective work. The best performances in each group are in \textbf{bold}.}
\label{tab:rg_result}
\end{table*}

\paragraph{Baseline} For intent classification, we compare with various few-shot fine-tuned baselines:  RoBERTa~\cite{liu2019roberta}, ICDA~\cite{lin2023icda}, DNNC~\cite{zhang2020dnnc}, and CPFT~\cite{zhang2021cpft}. While for zero-shot baseline, we employ MNLI~\cite{williams2018mnli} fine-tuned BART$_{large}$ models~\cite{lewis2020bart} by framing intent classification as an NLI task. For dialogue state tracking, we compare with multiple strong zero-shot baselines in the single-domain setting: TRADE \cite{wu2019transferable}, MA-DST \cite{kumar2020ma}, TransferQA \cite{lin2021zero} and T5Dep \cite{wang2022slot}. For the response generation, we compare with modular and non-unified end-to-end TODS---e.g., having split decoder modules for DST and response generation---including SFN~\cite{mehri2019sfn},  LAVA~\cite{lubis2020lava}, DAMD~\cite{zhang2020damd}, MARCO~\cite{wang2020marco}, MinTL~\cite{lin2020mintl}, HDSA~\cite{santra2021hdsa}, RSTOD~\cite{cholakov2022rstod}, and BORT~\cite{sun2022bort}.

\subsection{Key Takeaways}
\label{sec:insights}

\begin{figure*}[!t]
  \centering
  \includegraphics[width=0.88\linewidth]{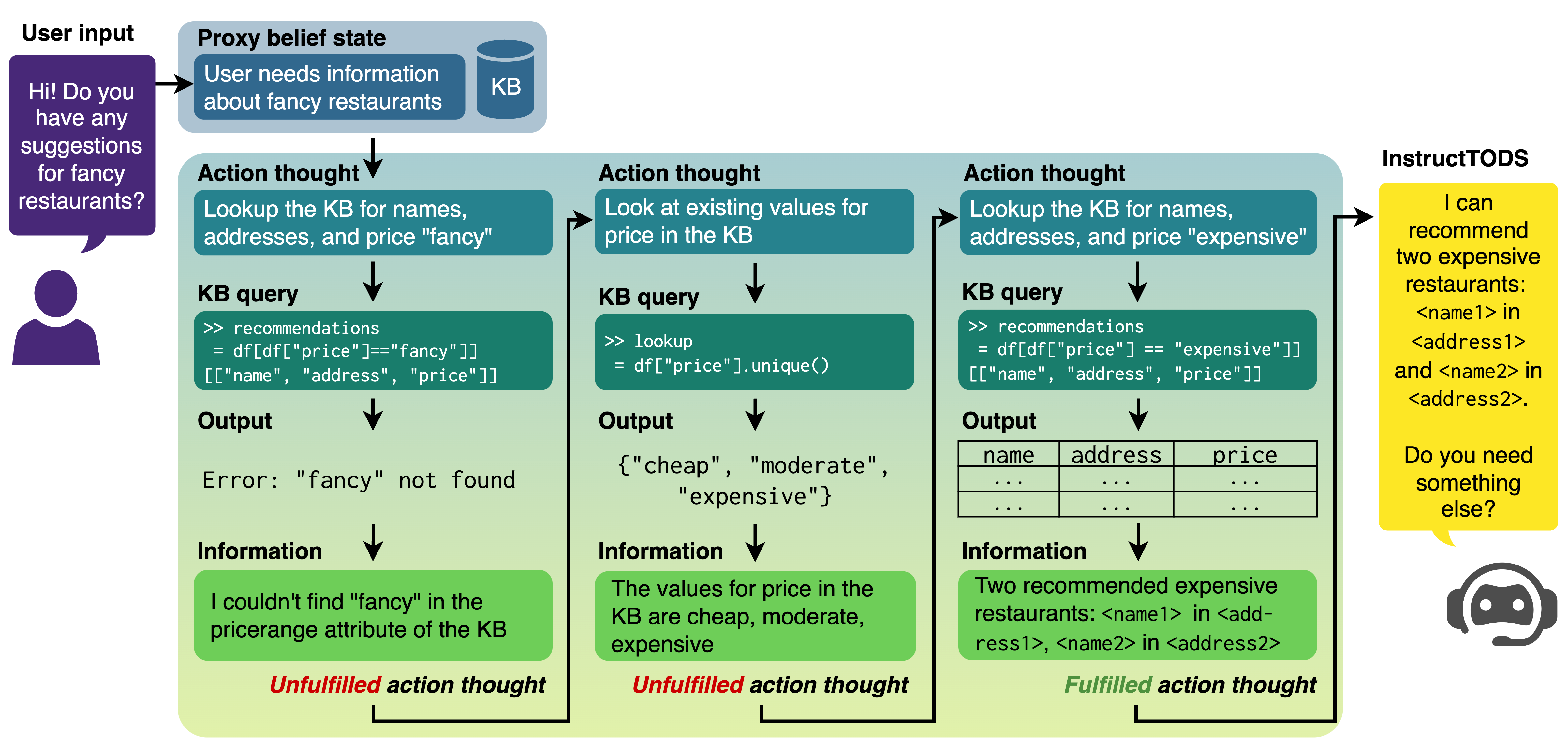}
  \caption{Overview of \textbf{InstructTODS}, a framework to utilize LLM for zero-shot end-to-end task-oriented dialogue.}
  \label{fig:e2e_framework}
\end{figure*}

The evaluation results for intent classification, DST, and response generation are shown in Table~\ref{tab:ic_results}, Table~\ref{tab:dst_result}, and Table~\ref{tab:rg_result}, respectively. We summarize the key insights as follows:

\paragraph{LLMs outperform most baselines.} 
LLMs show significant improvements in intent classification and DST tasks compared to other zero-shot and few-shot baselines and perform almost comparably to few-shot models in the intent classification task.

\paragraph{LLMs offer better generalization and adaptable solutions to TOD.} Unlike fine-tuned models, LLMs approach all tasks in an autoregressive generation manner, allowing greater flexibility and scalability to adapt to other tasks and domains.

\paragraph{LLMs generate responses that better reflect human preference.} Unlike other fine-tuned approaches, LLMs generate responses that are distinct from the gold responses resulting in lower BLEU scores. Nevertheless, the responses from modular LLMs are more fluent, diverse, and human-likable compared to baselines, while having competitive Inform and Success rates.

\paragraph{LLMs do not solve multi-domain DST problems}
Despite their strong performance, LLMs often over-predict active slots, leading to errors in the \textit{all slots} setting. LLMs tend to mix up parallel slots over different domains, especially for either temporal or spatial information, e.g., destination and departure, leave time and arrival time, etc.



\section{InstructTODS: An Instruction-Based Zero-shot End-to-End TODS}
\label{sec:method}

By leveraging the insights from solving TODS subtasks on \S\ref{sec:insights}, we develop the first zero-shot end-to-end framework that operates without any domain information (ontology) and requires no task-specific annotations such as dialogue state, system act, intent, etc. This method is not only cost-efficient but also alleviates the ontology constraint of LLMs in the modular DST task and promotes the strength of LLMs in generating better and more human-preferred responses. Let us define a structured knowledge base (KB) as a set of tuples $\mathcal{K} = \{(v^{a_1}_{1}, ..., v^{a_k}_{1}), ..., (v^{a_1}_{p}, ..., v^{a_k}_{p})\}$ where $(a_i)^{k}_{i=0}$ are the attributes of the KB, and $(v^{a_i}_j)^{p}_{j=0}$ are all the values associated to the attribute $a_i$. 

We first define a naive modular LLM response generation approach that serves as a baseline, denoted as \textbf{RG$_{naive}$}.\footnote{\textbf{RG$_{naive}$} is based on \citet{bang2023multitask}'s limited observation of zero-shot end-to-end TODS with ChatGPT (see \S\ref{sec:related-works-llms}).}
\textbf{RG$_{naive}$} generates the user response by taking the entire KB along with the dialogue context as input. In this approach, we rely on the ability of the LLM to parse the entire KB during inference while processing the dialogue context, in order to perform in-context retrieval and response generation at the same time. As such, we build the input $x^{RG}_i = \mathbb{P}^{RG}(\mathcal{I}^{RG}, f^{RG}(\mathcal{K}), \mathcal{D}_i)$ where $\mathbb{P}^{RG}(.)$ is the response generation input template, $\mathcal{I}^{RG}$ denotes the instruction for response generation and $f^{RG}(\mathcal{K})$ refers to a textual transformation of the KB where we filter unnecessary information and values that are too long as they are not needed to accomplish the user goal. In this approach, the bottleneck resides in the context window limit of the LLMs. 
Unlike other approaches, InstructTODS aims to make the best use of the LLM abilities to perform end-to-end tasks in zero-shot settings without the need for additional modular NLU and DST models, allowing zero-cost adaptation to various domains with no parameter update.

In general, in order to process the dialogue history and interact with the KB, InstructTODS introduces two concepts, i.e., proxy belief state and action thought. The results from KB and the dialogue history are then fed as a context to the LLM for generating the user response. 

In the following paragraphs, we describe each component of InstructTODS in more detail.

\paragraph{Proxy Belief State} We generate a proxy belief state $\tilde{B}_i = \mathbb{P}^{BS}(\mathcal{D}_i)$ from the dialogue history where $\mathbb{P}^{BS}(.)$ denotes the prompt template and $\mathcal{D}_i$ the dialogue context. $\tilde{B}_i$ encapsulates everything that the user is looking for in natural language at this point of the dialogue. Note that, the proxy belief state does not need any prior knowledge about the domain nor any ontology to operate (e.g. domain, trackable slots, values, types of information, etc.). The proxy belief state is directly used to interact with the KB in a multi-turn fashion.

\begin{table*}[!t]
\centering
\resizebox{0.92\linewidth}{!}{
    \begin{tabular}{l|cc|cc|cc|cc|cc|ccc}
    \toprule
     \multicolumn{1}{c}{\multirow{3}{*}{\textbf{Model}}} & \multicolumn{2}{|c}{Attraction} & \multicolumn{2}{|c}{Hotel} & \multicolumn{2}{|c}{Restaurant} & \multicolumn{2}{|c}{Taxi} & \multicolumn{2}{|c}{Train} & \multicolumn{3}{|c}{\textbf{All}}  \\
     & \textbf{Inf.} & \textbf{Succ.} & \textbf{Inf.} & \textbf{Succ.} & \textbf{Inf.} & \textbf{Succ.} & \textbf{Inf.} & \textbf{Succ.} & \textbf{Inf.} & \textbf{Succ.} & \textbf{BLEU} & \textbf{Inf.} & \textbf{Succ.} \\
     \midrule
    SOLOIST & \textbf{100} & \textbf{90.90} & \textbf{90.00} & \textbf{85.00} & 78.30 & 70.00 & \textbf{100} & \textbf{100} & 81.80 & 78.80 & 13.58 & 88.80 & 84.30 \\
    UBAR & \textbf{100} & \textbf{90.90} & 85.00 & 70.00 & 91.70 & 83.30 & \textbf{100} & 90.00 & 90.90 & 84.80 & 15.05 & 91.90 & 82.10 \\
    AUGPT & 90.90 & 81.80 & 71.70 & 60.00 & 81.70 & 73.30 & \textbf{100} & 84.00 & \textbf{97.00} & \textbf{93.90} & 15.56 & 86.10 & 76.20 \\
    GALAXY & 90.90 & 72.70 & 81.70 & 76.70 & 91.70 & 83.80 & \textbf{100} & \textbf{100} & 93.90 & \textbf{93.90} & \textbf{18.10} & 91.00 & \textbf{86.10} \\
    PPTOD & 81.80 & 81.80 & 71.70 & 71.70 & 86.70 & \textbf{86.70} & \textbf{100} & \textbf{100} & 89.20 & 84.80 & 16.44 & 89.20 & 84.80 \\
    \midrule
    RG$_{naive}$ & 81.80 & 36.37 & \textbf{90.00} & 83.33 & \textbf{96.70} & 83.33 & \textbf{100} & 89.80 & \textbf{97.00} & 63.67 & 3.95 & \textbf{94.90} & 82.16 \\
    \midrule
    InstructTODS & 72.70 & 54.55 & 85.00 & 75.00 & 91.70 & 73.33 & \textbf{100} & 89.80 & 90.90 & 72.73 & 3.94 & 90.70 & 76.20 \\
    \bottomrule
    \end{tabular}
}
\caption{\textbf{Task completion} performance comparison. InstructTODS have competitive Inform and Success rates compared to other end-to-end fine-tuned TODS baselines. \textbf{Bold} represents the highest score in each column.}
\label{tab:e2e_result}
\end{table*}

\paragraph{KB Interaction} To interact with the KB, we generate an Action thought $A = \mathbb{P}^{act}(\tilde{B}_i, (a_i)^{k}_{i=0})$ where $\mathbb{P}^{act}(.)$ is the template for action generation and $(a_i)^{k}_{i=0}$ the attributes of the KB. By providing the existing attributes of the KB at this step, we ground the LLM to accurately translate the belief state into information that can be queried from the KB, while filtering out unnecessary data. The action thought serves as an intermediary to leverage the code generation ability of LLM by generating a query $Q = \mathbb{P}^{KB}(A, \mathcal{K})$ where $\mathbb{P}^{KB}(.)$ is the template for code generation. The output from the KB is then parsed by the LLM to extract relevant information, denoted as $I$, presented in natural language, which provides a summary of the KB interaction. It also determines whether the current action thought has been fulfilled. If it remains unanswered, a new action thought is generated based on the extracted information, and the process repeats until a stopping criterion is reached indicating that no relevant knowledge is found in the KB.

\paragraph{Response Generation}  Once the KB interaction concludes, the final information, together with the original dialogue context, is passed to the model to generate the response $Y = \mathbb{P}^{RG}(I, \mathcal{D}_i)$ where $\mathbb{P}^{RG}$ represents the response generation template and $I$ the final information from the KB interaction. In the case where no knowledge is found in the KB, the LLM prompts the user to provide additional information. We provide the prompt template in Appendix \ref{sec:prompts}. The depiction of how the InstructTODS framework works is presented in Figure~\ref{fig:e2e_framework}.

\section{Experiment settings}

\paragraph{Baselines}

Our framework is compared to other end-to-end unified TODS approaches that perform end-to-end TODS using a unified text-to-text paradigm through a single generalized text generation model, i.e., SimpleTOD~\cite{ehsan2020simpletod}, PPTOD~\cite{su2022multi}, Soloist~\cite{peng2021soloist}, UBAR~\cite{yang2021ubar}, AuGPT~\cite{kulhanek2021augpt}, and Galaxy~\cite{he2022galaxy}. In addition, as described in \S\ref{sec:method}, we add the naive version of the LLM response generation approach which is fed by the full KB (\textbf{RG$_{naive}$}), as an additional baseline to better evaluate the effectiveness of our framework.
\begin{figure*}
    \centering
    \resizebox{0.85\linewidth}{!}{
        \includegraphics[width=0.4\linewidth, trim={3cm 0 0 0}]{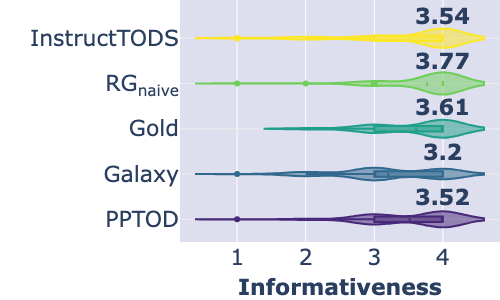}
        \includegraphics[width=0.29\linewidth]{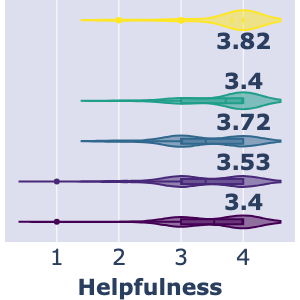}
        \includegraphics[width=0.29\linewidth]{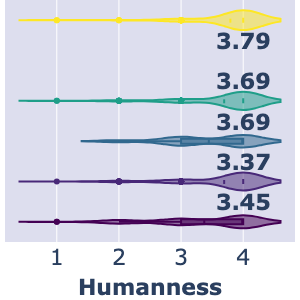}
    }

    \caption{Human evaluation comparison on informativeness \textbf{(left)}, helpfulness \textbf{(center)}, and humanness \textbf{(right)}. 
    }
    \label{fig:result_he}
\end{figure*}


\paragraph{Datasets}
We evaluate the end-to-end zero-shot capability on MultiWOZ 2.1 (MWOZ)~\cite{eric2020multiwoz,lewkowycz2022beyond}. We split the evaluation into two settings, i.e., single-domain and multi-domain evaluation settings, where we show the capability of LLMs to tackle more complex TODS tasks in zero-shot end-to-end settings.

\paragraph{Automatic Evaluation}

For evaluating the end-to-end framework,
we measure the per domain Inform and Success rates, and the BLEU~\cite{papineni2002bleu}, Inform rate, and Success rate~\cite{eric2020multiwoz} for all domains. The evaluation metric is computed on the delexicalized responses to avoid favoring models that provide more information than others and focus solely on the vocabulary used for the response generation. Additionally, we also incorporate an automatic human-likability score, namely USL-H~\cite{phy2020uslh}.


\paragraph{Human Evaluation}
We conduct an extensive human evaluation to measure the capability of LLMs in conducting zero-shot end-to-end unified TOD. Specifically, we conduct two human evaluations, which measure: 1) the informativeness, helpfulness, and humanness of the generated responses, and 2) the information correctness and hallucination rate of our InstructTODS. For evaluating informativeness, helpfulness, and humanness, we ask 3 annotators to rate the quality of the response using a 4-point Likert scale (see Appendix \ref{sec:human_eval_details}). The system is helpful if it answers the user's request while pushing the conversation towards goal completion, informative if the system provides enough related information while answering the user, and human if the generated answer is fluent and human-preferred. 
For measuring the incorrectness and the hallucination rate, the metrics are evaluated by a single TOD expert. The incorrectness and hallucination rate are measured by manually checking the ratio of correct, incorrect, and hallucinated entities provided in the generated responses. We conduct the human evaluation by taking 50 generated responses from all the models and the gold responses.


\section{Results and Analysis}
\label{sec:results}

\subsection{Automatic Evaluation}
Our automatic evaluation is shown in Table~\ref{tab:e2e_result}. In general, we find a similar trend with the modular LLMs where LLMs produce lower BLEU scores---$\sim$4 BLEU against $\sim$15 BLEU---with competitive Inform and Success rates compared to other end-to-end unified TODS baselines. Note that, as mentioned in \S\ref{sec:insights}, LLMs often generate completely different responses to the gold knowledge, hence producing low automatic evaluation scores. Nevertheless, the low automatic evaluation scores do not sufficiently reflect the capability of InstructTODS. We will further elaborate on this in \S\ref{sec:human-eval}, raising a question of the sufficiency of evaluating TODS quality using only a single gold response. Some comparative generation samples between the different models can be found in Appendix \ref{sec:appendix_samples}.

\begin{figure}[t]
    \centering
    \resizebox{0.85\linewidth}{!}{
    \includegraphics[trim={3cm 0 0 0}]{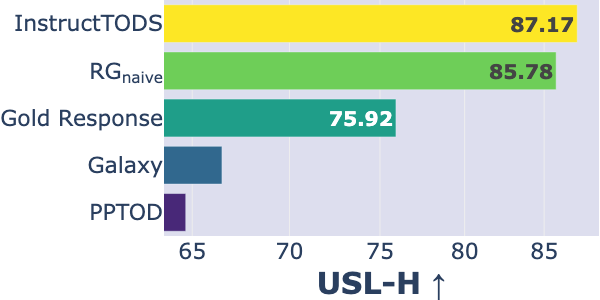}}
    \caption{InstructTODS have higher human preference scores than the gold responses and baselines.}
    \label{fig:e2e-usl-h}
\end{figure}

\subsection{Informativeness, Helpfulness, and Humanness of InstructTODS}
\label{sec:human-eval}

The results for our human evaluation are shown in Figure \ref{fig:result_he} for InstructTODS in comparison with the naive approach, the gold responses, and the two best-performing baselines in task completion (i.e., Galaxy and PPTOD). From the results, we show that InstructTODS is more informative, helpful, and human-like than the two fine-tuned end-to-end baselines by a noticeable margin. For both helpfulness and humanness, InstructTODS also outperforms RG$_{naive}$ and the gold response. Aligning with the human evaluation results, the generated responses by our framework also have higher humanness scores as shown in Figure \ref{fig:e2e-usl-h}, even higher than the gold responses. RG$_{naive}$ is the most informative, which is expected as the model processes the entire KB for information, however, the quality of the information greatly differs as shown in\S\ref{sec:incorrect_halucinate}.


\subsection{Incorrectness and Hallucination}
\label{sec:incorrect_halucinate}

\begin{figure}[t]
    \centering
    \includegraphics[width=0.95\linewidth, trim={0 0 2cm 0}]{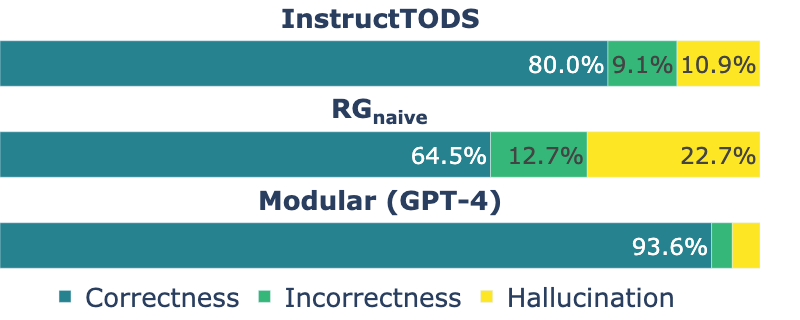}
    \caption{Human evaluation on correctness, incorrectness, and hallucination for \textbf{RG$_{naive}$} and \textbf{InstructTODS}.}
    \label{fig:eval-correctness}
\end{figure}

We show the results for incorrectness and hallucination for the LLM-generated responses in Figure~\ref{fig:eval-correctness}. While a sample can be incorrect, e.g., if the LLM database interaction fails, the LLMs do not necessarily generate unfaithful information. InstructTODS is more robust than naively employing the LLMs, improving the correctness by 15\% and showing 11\% of hallucination, half the amount of the RG$_{naive}$. We observe that some types of information are more prone to hallucination, notably time and address. This bias towards temporal and spatial information aligns with our observation of LLMs' performance in DST (\S\ref{sec:insights}).


\subsection{LLMs on Multi-Domain TOD}

\begin{figure*}[!t]
  \centering
  \resizebox{0.9\linewidth}{!}{
      \includegraphics[width=0.3\linewidth]{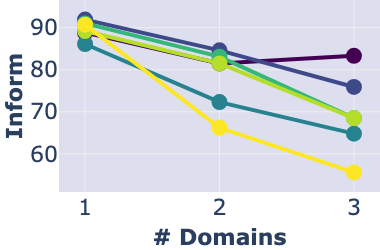}
      \hspace{1.5em}
      \includegraphics[width=0.475\linewidth]{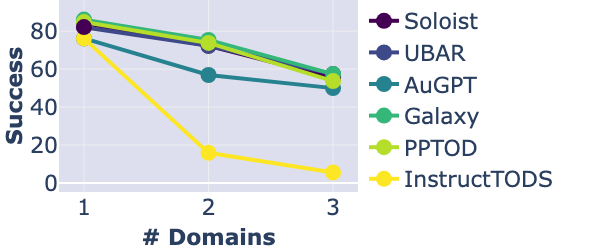}
  }
  \caption{End-to-end TODS performance degrades as the number of active domains in the dialogue increases.}
  \label{fig:e2e-multi-domain}
\end{figure*}

While it is possible to use InstructTODS in multi-domain with distinct KBs per domain, as we see in Figure \ref{fig:e2e-multi-domain}, the performance degrades quickly for Success and slightly less for Inform as the number of domains increases. While fine-tuned end-to-end baselines operate with only one KB at a given turn by tracking the active domain through either state changes \cite{peng2021soloist,yang2021ubar} or slot names \cite{kulhanek2021augpt}, our zero-shot framework does not assume any external knowledge nor ontology information. As such, all KBs are provided at each turn, and due to different KBs attributes overlapping in MWOZ, InstructTODS often queries incompatible information from the proxy belief state (e.g., "food" and "destination" at the same time), which are in different KBs. Hence, multi-domain degradation is largely due to the KB interaction failure.





\section{Related Work}



\subsection{Task-Oriented Dialogue System}

Task-oriented dialogue systems (TODS) can be broadly classified into two categories~\cite{chen2017todsurvey,gao2018convai,zhang2020todsurvey} which include pipelined approaches~\cite{nagata1994first,levin1997learning,levin2000stochastic,hurtado2005stochastic,williams2007partially,hori2009statistical,lee2009example} and end-to-end approaches~\cite{madotto2018mem2seq,wu2019global,madotto2020attention,raghu2019disentangling,qin2020dynamic,hosseini2020simple,lei2018sequicity,ehsan2020simpletod,lin2020mintl,he2022galaxy,kulhanek2021augpt,peng2021soloist,su2022multi,yang2021ubar}. The pipelined approach utilizes multiple modules in order to generate the system responses. While the end-to-end approach directly generates responses from the user input and the KB in an end-to-end manner.

\paragraph{End-to-end TODS} Early approaches for end-to-end TODS employ template responses in a retrieval or generation setting \cite{zhao2017generative,eric2017key,wu2019global}.
While other approaches inject KBs directly into the model to perform end-to-end generation \cite{madotto2018mem2seq,madotto2020learning}. 
A more recent end-to-end TODS tackles end-to-end response generation in a single sequence prediction problem \cite{hosseini2020simple,yang2021ubar,peng2021soloist} with an autoregressive model. These approaches still mostly leverage TOD data (belief states, system acts, etc.) during generation. As general pre-trained LMs were shown to be effective for TODS \cite{mehri2019sfn,lubis2020lava,lin2020mintl}, several subsequent works have explored pre-training approaches directly tailored towards TODS~\cite{zhang2020damd,su2022multi,he2022galaxy}. To the best of our knowledge, prior works require a structured format of dialogue states, system acts, and/or template responses, whereas InstructTODS alleviates such needs by incorporating an unstructured proxy belief state, which requires no domain-specific knowledge nor ontology to operate, allowing zero-shot adaptation to various TOD domains. 

\subsection{Zero-Shot Generalization of LLMs}
\label{sec:related-works-llms}


LLMs have shown
remarkable zero-shot generalization capabilities in various NLP tasks~\cite{brown2020language,scao2022bloom,chowdhery2022palm,thoppilan2022lamda}. This is further improved through instruction tuning~\cite{wei2021finetuned,sanh2021multitask,wei2022chain,chung2022scaling,longpre2023flan,cahyawijaya2023instructalign}, which enables a better generalization to unseen tasks, and reinforcement learning with human feedback~\cite{christiano2015rlhf,ouyang2022training,bai2022training}, which enables a better alignment of human preferences. The zero-shot generalization ability of LLMs has also been explored in more specific cases, e.g., multiple choice question answering~\cite{robinson2023leveraging}, biomedical NLP~\cite{fries2022bigbio}, reasoning~\cite{bang2023multitask}, low-resource languages~\cite{cahyawijaya2023nusacrowd,cahyawijaya2023instructalign,asai2023buffet}, code-switching~\cite{yong2023prompting,zhang2023multilingual}.
\paragraph{LLMs for TODS}
Recent works explore the applicability of LLMs in solving modular TOD tasks~\cite{bang2023multitask,hudevcek2023llms} and a pipeline manner~\cite{ehsan2020simpletod, su2022multi, peng2021soloist, yang2021ubar, kulhanek2021augpt, he2022galaxy}. Additionally, \citet{bang2023multitask} inspect ChatGPT's capability for zero-shot end-to-end TODS, however, it is limited to only $\sim$1\% of the test set available. Therefore, to the best of our knowledge, our work is the first to comprehensively study the utilization of LLMs for zero-shot end-to-end TODS.

\section{Conclusion}
In this paper, we introduce InstructTODS, an off-the-shelf framework to effectively perform end-to-end TODS in zero-shot utilizing LLMs. We compare InstructTODS to several state-of-the-art fully fine-tuned end-to-end TODS and show that InstructTODS manages to guide the conversation towards goal completion similarly to the fine-tuned systems on MWOZ while generating answers that are more informative, helpful, and human-like than previous approaches. Furthermore, we investigate the capability of LLMs in performing various TOD subtasks in zero-shot settings, demonstrating better diversity and human preference on response generation, and state-of-the-art zero-shot results on dialogue state tracking and intent classification. 



\section{Limitation}

\paragraph{Generalization to Other Datasets}
In our work, we only assess the effectiveness of InstructTODS on MultiWoZ 2.1 dataset, whose size is a magnitude higher than other TODS datasets~\cite{eric2020multiwoz}. We conjecture that the generalization to other datasets will follow the same trend as described in \S\ref{sec:results}, where it excels in the single-domain setting while still struggling in the multi-domain setting. We expect future work to extend the assessment on InstructTODS to other datasets and domains.

\paragraph{Generalization to Other Languages}
In recent years, various task-oriented dialogue systems in languages other than English have been introduced, such as CrossWoZ~\cite{zhu2020crosswoz}, BiTOD~\cite{lin2021bitod}, GlobalWoZ~\cite{ding2022globalwoz}, and COD~\cite{majewska2023cross}. As suggested in prior works evaluating LLMs in low-resource languages~\cite{bang2023multitask,asai2023buffet,cahyawijaya2023instructalign,cahyawijaya2023nusacrowd,workshop2023bloom,kabra2023multilingual,zhang2023multilingual}, we conjecture that the performance in other languages follow the general trend in LLMs where the performance in low-resource languages will be lower compared to the high-resource languages. Future work might explore and further extend methods for improving the generalization of InstructTODS to other languages.

\paragraph{Generalization to Other LLMs} In this work, we only explore two proprietary LLMs which display strong performance on various NLP tasks, i.e., GPT-3.5 and GPT-4. Despite the lack of transparency of these models, we expect that when other publicly available LLMs achieve the same performance as these proprietary LLMs, a similar capability of zero-shot end-to-end TODS will emerge. We expect future work to explore the generalization of InstructTODS and its improvement in other LLMs.


\section{Ethics Statement}


Our research endeavors to develop an off-the-shelf framework for zero-shot end-to-end Task-Oriented Dialogue Systems (TODS) using Large Language Models (LLMs). It is important to note that this study does not involve the use of any sensitive data and the experimental evaluation is conducted on publicly available datasets. To ensure the quality of our results, we have employed crowdsourcing for the human evaluation of the generated dialogue responses. While our study does not raise any ethical concerns regarding privacy, confidentiality, or bias, we recognize that the use of LLMs in dialogue systems may have ethical implications related to potential biases in the training data and the generated responses. Therefore, we emphasize the importance of ongoing research toward developing ethical guidelines and best practices for the use of LLMs in dialogue systems. In line with our commitment to transparency and reproducibility, we will be releasing our code publicly. We believe that this will encourage open and collaborative research towards the development of more ethical and effective dialogue systems.

\newpage
\newpage

\bibliography{anthology,custom}
\bibliographystyle{acl_natbib}

\newpage

\appendix
\label{sec:appendix}




\setcounter{table}{0}
\renewcommand{\thetable}{A\arabic{table}}
\setcounter{figure}{0}
\renewcommand{\thefigure}{A\arabic{figure}}

\section{Comparison of LLMs over Various NLP Tasks}
\label{app:llm-comparison}

\begin{figure}[!h]
    \centering
    \includegraphics[width=\linewidth]{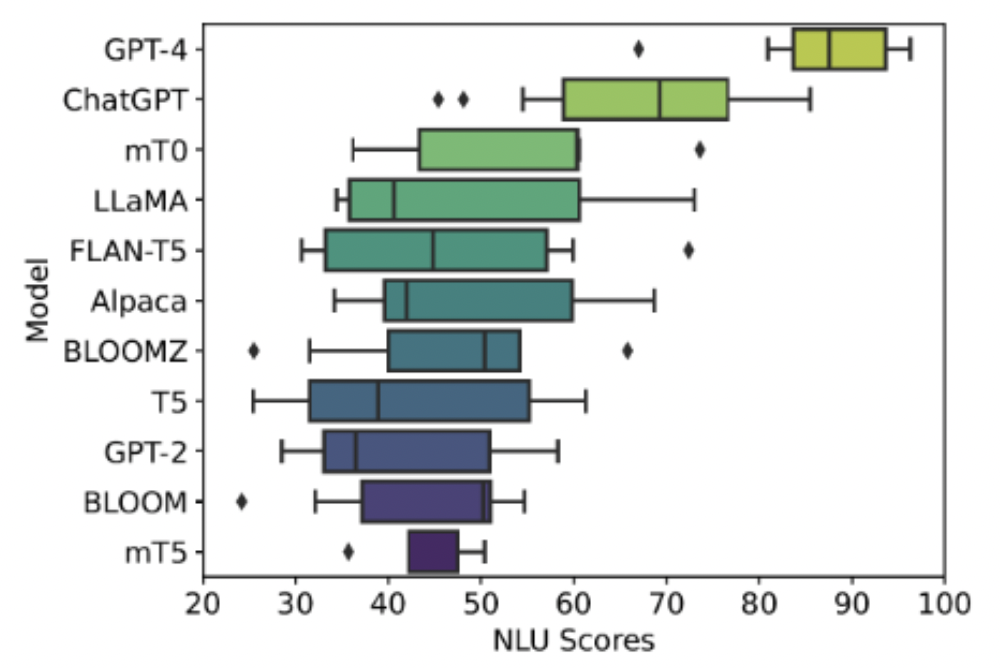}
    \includegraphics[width=\linewidth]{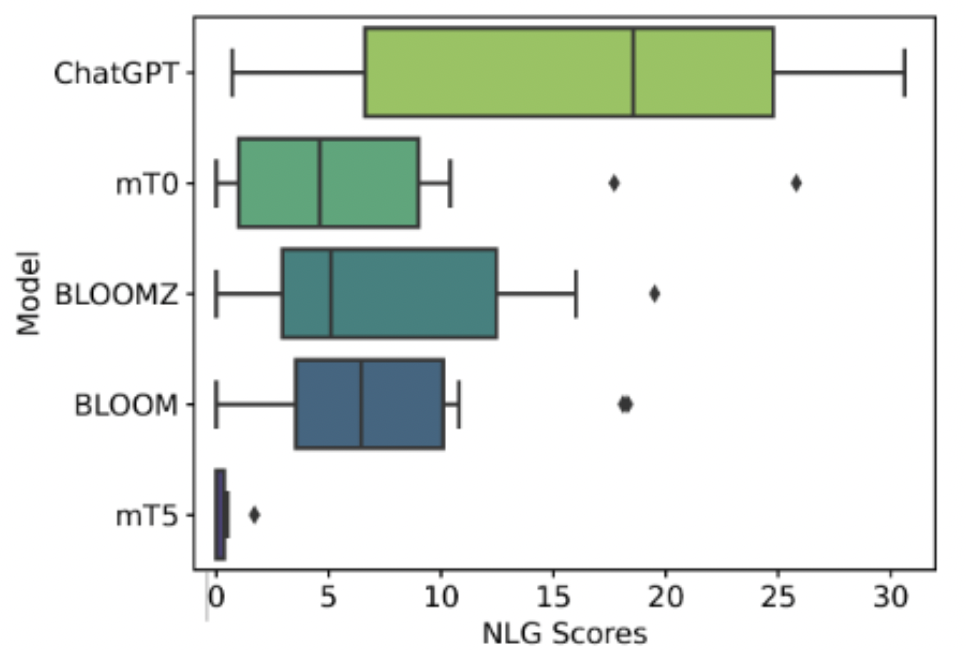}
    \caption{Average LLM performance comparison on various \textbf{(left)} NLG and \textbf{(right)} NLU tasks.}
    \label{fig:llm_comparison}
\end{figure}

We show the performance comparison of various LLMs on both NLU and NLG tasks in Figure~\ref{fig:llm_comparison}. The data are collected from various prior works focusing on benchmarking the capabilities of LLMs~\cite{bang2023multitask,cahyawijaya2023nusacrowd,anonymous2023nusawrites,asai2023buffet,openai2023gpt4,wu2023laminilm}.

\section{Human Evaluation}
\label{sec:human_eval_details}
We give additional details concerning the human evaluation in this section. The instructions for each metric given to the evaluators are defined as follow:

\paragraph{Informativeness} Amount of information that the system provides while answering the user’s utterance.
\begin{enumerate}
    \item The response has no information at all
    \item The response provides at least one piece of information, but clearly not enough.
    \item The response provides several pieces of information but more could be provided
    \item The response gives all the information you would expect in that turn
\end{enumerate}

\paragraph{Helpfulness} The system answers the user's utterance and pushes the conversation towards completion
\begin{enumerate}
    \item The response is doing neither
    \item The response is just pushing the conversation towards completion without answering the question
    \item The response is just answering the question but not pushing the conversation towards completion
    \item The response is doing both
\end{enumerate}

\paragraph{Humanness} The system answers the user's utterance in a human-like manner
\begin{enumerate}
    \item The response is completely machine-like
    \item The response contains human-like acknowledgement, but still sound machine-like
    \item The response has human acknowledgement and you could imagine it being said by a human
    \item The response is completely indiscernible from a human answer
\end{enumerate}

\section{Generation Samples}
\label{sec:appendix_samples}

We show a few generation samples of InstructTODS as well as the modular evaluation of TODS subtasks. In table \ref{tab:e2e_sample}, we show different delexicalized responses of the fine-tuned baseline compared to the generation by InstructTODS. For the modular subtasks, we show a sample of correct prediction as well as difference failure cases. The samples for dialogue state tracking are in Table \ref{tab:DST_samples}, for intent classification in Table \ref{tab:IC_samples} and for response generation in Table \ref{tab:RG_samples}.

\begin{table*}[t!]
\centering
\footnotesize
\begin{tabular}{p{0.10\linewidth} | p{0.89\linewidth}}
\toprule
ID & SNG0991.json\\ 
\midrule
Dialogue Context & Alright, let's try for a cheap hotel.\\
\midrule
Gold & Do you want a cheap hotel or a cheap guesthouse?\\
\midrule
SOLIST & I have [value\_count] [value\_pricerange] hotels. [value\_count] in the [value\_area] and [value\_count] in the [value\_area]. Which area do you prefer ?\\
\midrule
UBAR & What area would you like to stay in?\\
\midrule
AUGPT & I have [value\_count] [value\_price range] [value\_type] available. Do you have a preference for the area?\\
\midrule
GALAXY & [value\_name] meets your criteria. Would you like me to book you a room ?\\
\midrule
PPTOD & There are [value\_choice] [value\_price] [value\_type] with free parking. Do you have a preference for the area ?\\
\midrule
InstrucTODS (Zero-Shot) & Great, I found [value\_name] located at [value\_address] [value\_postcode]. It has free parking and falls within the [value\_pricerange] price range. Would you like me to confirm a booking for you?\\
\midrule[\heavyrulewidth]
\midrule[\heavyrulewidth]
ID & MUL2320.json\\ 
\midrule
Dialogue Context & No, that's everything I needed today. Thank you!\\
\midrule
Gold & Ok, thank you for calling. Goodbye.\\
\midrule
SOLIST & You are welcome. Have a great day !\\
\midrule
UBAR & Thank you for using our system!\\
\midrule
AUGPT & You're welcome. Have a great day!\\
\midrule
GALAXY & You are welcome. Have a great day !\\
\midrule
PPTOD & You are welcome. Have a great day !\\
\midrule
InstrucTODS (Zero-Shot) & Great, let me know if you need any further assistance. Have a great day!\\
\midrule[\heavyrulewidth]
\midrule[\heavyrulewidth]
ID & MUL1489.json\\ 
\midrule
Dialogue Context & USER: Yes, please. I need a table for 8 on Friday at 17:30, please.

SYSTEM: Ok, you are all set for cote on Friday, table for 8 at 17:30. Can I help with anything else?

USER: Can I have the reference number for the reservation please?

SYSTEM: Booking was unsuccessful. Can you try another time slot?

USER: What about 16:30? \\
\midrule
Gold & One moment while I try to make the reservation of table for 8, Friday at 16:30.\\
\midrule
SOLIST & I have made that reservation for you. Your reference number is [value\_reference]. Is there anything else I can help you with ?\\
\midrule
UBAR & I was able to book your table for [value\_day] at [value\_time]. Your reference number is [value\_reference].\\
\midrule
AUGPT & Yes, the reference number is [value\_reference].\\
\midrule
GALAXY & I am sorry, but I was unable to book that for you. Would you like to find another restaurant?\\
\midrule
PPTOD & Your reference number is [value\_reference]. Is there anything else I can help you with ?\\
\midrule
InstrucTODS (Zero-Shot) & Sure your reference number for the reservation at [value\_name] on [value\_day] at [value\_time] for a table of [value\_people] is [value\_reference]. Is there anything else I can assist you with?\\
\bottomrule
\end{tabular}
\caption{Samples of different system responses for both the proposed E2E framework in comparison with the gold response. InstrucTOD's response are delexicalized after generation, while the fine-tuned systems generate the delexicalized response directly.}
\label{tab:e2e_sample}
\end{table*}

\begin{table*}[t!]
\centering
\footnotesize
\begin{tabular}{p{0.10\linewidth} | p{0.89\linewidth}}
\toprule
ID & SNG01679.json\\ 
\midrule
Prompt & Generate the dialogue state of the following dialogue between a USER and a task-oriented dialogue SYSTEM. The results should be in a single python dictionary following this format: {"domain1-slot1":value1, "domain2-slot2":"value2"}. Use the provided domain and slots, and nothing else: \newline \newline
SLOTS: \newline
attraction-area, attraction-name, [...], hotel-day, hotel-stars, [...], train-leaveAt \newline
\newline
CONTEXT: \newline
USER: I need a taxi to take me to Pipasha Restaurant to leave after 01:45. \newline \newline
BELIEF STATES:
\\
\midrule
Gold & \{'taxi-leaveAt': '01:45', 'taxi-destination': 'pipasha restaurant'\}
\\
\midrule
Modular (GPT-4) & \{"taxi-leaveAt":"01:45", "taxi-destination":"Pipasha Restaurant"\}
\\
\midrule[\heavyrulewidth]
\midrule[\heavyrulewidth]
ID & SNG01936.json\\ 
\midrule
Prompt & Generate the dialogue state of the following dialogue between a USER and a task-oriented dialogue SYSTEM. The results should be in a single python dictionary following this format: {"domain1-slot1":value1, "domain2-slot2":"value2"}. Use the provided domain and slots, and nothing else: \newline \newline
SLOTS: \newline
attraction-area, attraction-name, [...], hotel-day, hotel-stars, [...], train-leaveAt \newline
\newline
CONTEXT: \newline
USER: I am looking for city centre north b and b \newline
SYSTEM: I have found the guesthouse you were wanting. Would you like me to book this for you? \newline
USER: Yes, please book it for 1 person and for 5 nights starting  Friday. \newline \newline
BELIEF STATES:
\\
\midrule
Gold & \{'hotel-stay': '5', 'hotel-day': 'friday', 'hotel-people': '1', 'hotel-name': 'city centre north b and b'\}
\\
\midrule
Modular (GPT-4) & \{"hotel-stay": "5", "hotel-day": "Friday", "hotel-people": "1", \textcolor{red}{"hotel-area": "city centre north"},  \textcolor{red}{"hotel-type": "b and b"}\}
\\ 	
\midrule
Comment & In this example, the LLM mistakes the first part of the name "city centre north b and b" as "hotel-area" since it contains "centre north", and the end of the name as a "hotel-type" of hotel "b and b". \\
\midrule[\heavyrulewidth]
\midrule[\heavyrulewidth]
ID & SNG0263.json\\ 
\midrule
Prompt & Generate the dialogue state of the following dialogue between a USER and a task-oriented dialogue SYSTEM. The results should be in a single python dictionary following this format: {"domain1-slot1":value1, "domain2-slot2":"value2"}. Use the provided domain and slots, and nothing else: \newline \newline
SLOTS: \newline
attraction-area, attraction-name, [...], hotel-day, hotel-stars, [...], train-leaveAt \newline
\newline
CONTEXT: \newline
USER: I am looking for a train from Cambridge to Birmingham New Street. \newline
SYSTEM: The next train leaving from Cambridge for Birmingham New Street departs Friday at 5:01, and will arrive by 7:44. \newline
USER: I need a train that departs after 08:30 on Friday. \newline
SYSTEM: I have a train leaving Cambridge arriving at Birmingham New Street on Friday at \textcolor{red}{9:01}.  Would you like me to book this for you? \newline
USER: Yes please, that sounds perfect. \newline \newline
BELIEF STATES:
\\
\midrule
Gold & \{'train-leaveAt': '08:30', 'train-destination': 'birmingham new street', 'train-day': 'friday', 'train-departure': 'cambridge'\}
\\
\midrule
Modular (GPT-4) & \{"train-leaveAt":"08:30", "train-destination":"Birmingham New Street", "train-day":"Friday", "train-departure":"Cambridge", \textcolor{red}{"train-arriveBy":"9:01"}\}
\\ 	
\midrule
Comment & In this example, the LLM mistakenly parse the time given in the system utterance as "train-arriveBy". This type of mistake happens often in domains with temporal information as the dialogue history gets longer \\
\bottomrule
\end{tabular}
\caption{Samples of Zero-Shot Dialogue State Tracking. First example is correct, the other two show different type of LLM mistakes.}
\label{tab:DST_samples}
\end{table*}

\begin{table*}[t!]
\centering
\footnotesize
\begin{tabular}{p{0.10\linewidth} | p{0.89\linewidth}}
\toprule
Prompt & Select the correct intent that is expressed in the given SENTENCE among the list of INTENTS provided. Generate the correct intent and nothing more: \newline \newline
INTENTS: \newline
0 activate\_my\_card \newline
1 age\_limit \newline
2 apple\_pay\_or\_google\_pay \newline
[...]  \newline
74 why\_verify\_identity \newline
75 wrong\_amount\_of\_cash\_received \newline
76 wrong\_exchange\_rate\_for\_cash\_withdrawal \newline
 \newline
SENTENCE: \newline
"Is there a way to know when my card will arrive?"
\\
\midrule
Gold & card\_arrival
\\
\midrule
Modular (GPT-4) & 11 card\_arrival
\\
\midrule[\heavyrulewidth]
\midrule[\heavyrulewidth]
Prompt & Select the correct intent that is expressed in the given SENTENCE among the list of INTENTS provided. Generate the correct intent and nothing more: \newline \newline
INTENTS: \newline
0 activate\_my\_card \newline
1 age\_limit \newline
2 apple\_pay\_or\_google\_pay \newline
[...] \newline
74 why\_verify\_identity \newline
75 wrong\_amount\_of\_cash\_received \newline
76 wrong\_exchange\_rate\_for\_cash\_withdrawal \newline
 \newline
SENTENCE: \newline
"Can I reactivate my \textcolor{red}{lost} card that I found this morning in my jacket pocket?"
\\
\midrule
Gold & card\_linking
\\
\midrule
Modular (GPT-4) & \textcolor{red}{41 lost\_or\_stolen\_card}
\\
\midrule
Comment & The LLM focuses more on "lost" in the user utterance, although the emphasis to understand the correct intent should be put on "reactivate". \\
\midrule[\heavyrulewidth]
\midrule[\heavyrulewidth]
Prompt & Select the correct intent that is expressed in the given SENTENCE among the list of INTENTS provided. Generate the correct intent and nothing more: \newline \newline
INTENTS: \newline
0 activate\_my\_card \newline
1 age\_limit \newline
2 apple\_pay\_or\_google\_pay \newline
[...] \newline
74 why\_verify\_identity \newline
75 wrong\_amount\_of\_cash\_received \newline
76 wrong\_exchange\_rate\_for\_cash\_withdrawal \newline
 \newline
SENTENCE: \newline
"Why won't my card show up on the app?"
\\
\midrule
Gold & card\_linking
\\
\midrule
Modular (GPT-4) & \textcolor{red}{card\_not\_working}
\\
\midrule
Comment & In the utterance context, the main source of confusion is due to the card linking being implicit (by mentionning the app), while the card not working is more explicit (not show up).\\
\bottomrule
\end{tabular}
\caption{Samples of Zero-Shot Intent Classification. First example is correct, the other two show different type of LLM mistakes.}
\label{tab:IC_samples}
\end{table*}

\begin{table*}[t!]
\centering
\footnotesize
\begin{tabular}{p{0.10\linewidth} | p{0.89\linewidth}}
\toprule
ID & PMUL3027.json\\ 
\midrule
Prompt & In a task oriented dialogue setting, generate a SYSTEM response to the USER query in the conversation provided in CONTEXT. You should follow the information provided in ACT to generate this answer. Do not answer with anything other than what is provided in the dialogue act: \newline \newline
CONTEXT: \newline
USER: I am staying in the centre of town for the weekend, what is there to do there? \newline
SYSTEM: We have several things to do! Architecture, colleges, museums...What type of attraction are you most interested in? \newline
USER: It doesn't matter but can you recommend one and give me the entrance fee? \newline
ACT:Recommend the user for free, castle galleries. \newline
SYSTEM: 
\\
\midrule
Gold & I recommend castle galleries and it's free to get in!
\\
\midrule
Modular (GPT-4) & I recommend visiting the castle galleries, which are a great attraction and the best part is that the entrance is free!
\\
\midrule[\heavyrulewidth]
\midrule[\heavyrulewidth]
ID & PMUL3027.json\\ 
\midrule
Prompt & In a task oriented dialogue setting, generate a SYSTEM response to the USER query in the conversation provided in CONTEXT. You should follow the information provided in ACT to generate this answer. Do not answer with anything other than what is provided in the dialogue act: \newline \newline
CONTEXT: \newline
USER: I want to find a moderately priced restaurant. \newline
SYSTEM: I have many options available for you! Is there a certain area or cuisine that interests you? \newline
USER: Yes I would like the restaurant to be located in the center of the attractions.  \newline
SYSTEM: There are 21 restaurants available in the centre of town. How about a specific type of cuisine? \newline
USER: i need to know the food type and postcode and it should also have mutliple sports \newline
SYSTEM: I am sorry I do not understand what you just said. Please repeat in a way that makes sense.  \newline
USER: get me the food type and the post code \newline
SYSTEM: My favorite it The Copper Kettle at 4 Kings Parade City Centre CB21SJ.  It serves British food.  Does that interest you? \newline
USER: Absolutely! Thank you! I also need information on the attractions that have multiple sports in town, in the same area as the restaurant please.  \newline
SYSTEM: The only multiple sports attraction is located in the east of town. Would you like more information? \newline
USER: No, I want to be in the centre of town. What about architecture attractions? \newline
SYSTEM: All Saints Church is an architectural attraction. It is in the Centre of town and entrance is free! \newline
USER: Can I get their phone number and address please? \newline
SYSTEM: Their phone number is 01223452587. Is there anything else I can assist you with? \newline
USER: Oh, what type of architecture is it? \newline
ACT:Inform the user that the type is architecture. \newline
SYSTEM:
\\
\midrule
Gold & It is architecture type. Anything else?
\\
\midrule
Modular (GPT-4) & All Saints Church features \textcolor{red}{Gothic} architecture.
\\
\midrule
Comment & The type of architecture "Gothic" is hallucinated by the LLM here. \\
\bottomrule
\end{tabular}
\caption{Samples of Zero-Shot Response Generation using oracle system action. Second example show hallucination.}
\label{tab:RG_samples}
\end{table*}

\section{Prompts}
\label{sec:prompts}

In table \ref{tab:prompt_templates}, we give the templates and samples of the prompts used in $RG_{naive}$, and in the end-to-end setting for the proxy belief state, the initial knowledge base interaction and the response generation using the extracted information.

\begin{table*}[t!]
\centering

\footnotesize
\begin{tabular}{p{0.10\linewidth} | p{0.44\linewidth} | p{0.45\linewidth}}
\toprule
\textbf{Setting} & \textbf{Template} & \textbf{Sample} \\
\midrule
$RG_{naive}$ & Generate the answer of the SYSTEM in the following conversation between a USER and a SYSTEM in a task-oriented dialogue setting. You can either request more details to the user that is available in the knowledge base to complete the goal, or simply answer the user's request. Do not provide multiple choice for the user to choose, just recommend one, and generate nothing other than the SYSTEM reply. Use the following knowledge base to interact with the user and perform \{TASK\}: \newline \newline
\{DATABASE\} \newline \newline
\{DIALOGUE\_CONTEXT\} \newline
SYSTEM: \newline
 & Generate the answer of the SYSTEM in the following conversation between a USER and a SYSTEM in a task-oriented dialogue setting. You can either request more details to the user that is available in the knowledge base to complete the goal, or simply answer the user's request. Do not provide multiple choice for the user to choose, just recommend one, and generate nothing other than the SYSTEM reply. Use the following knowledge base to interact with the user and perform restaurant booking: \newline \newline
name: ..., address: ..., food: ..., \newline
name: ..., address:..., food: ..., \newline \newline
USER: Hi, any indian restaurants here? \newline
SYSTEM: Yes, we have Indian Express in the intermediate pricerange, would you want to know more? \newline
USER: Do you have any restaurant in the cheap pricerange? \newline
SYSTEM: \newline \\

\midrule
Proxy BS & Suppose you have access to a database with all necessary INFORMATION, what do you need to query to the database in order to reply to the user in the following conversation? \newline 
INFORMATION: \{SLOTS\} \newline \newline
You can follow these examples: \{EXAMPLES\} \newline \newline
\{DIALOGUE\_CONTEXT\} \newline
Need:
 & Suppose you have access to a database with all necessary INFORMATION, what do you need to query to the database in order to reply to the user in the following conversation? \newline
INFORMATION: address, area, name, phone, postcode, pricerange, entrance fee, food, internet, parking, stars \newline \newline
You can follow these examples: 
\newline
USER: I need fruits. \newline
SYSTEM: Do you have any preferences? \newline
USER: Yes, apples if possible. How expensive are they? \newline
Need: Information about pricerange for apple\newline \newline

USER: Hi, any indian restaurants here? \newline
SYSTEM: Yes, we have Indian Express in the intermediate pricerange, would you want to know more? \newline
USER: Do you have any restaurant in the cheap pricerange? \newline
Need:
 \\
 
\midrule
Initial KB Interaction & If there are multiple options fitting this criteria, pick a few to propose: \{Proxy BS\} & If there are multiple options fitting this criteria, pick a few to propose: Information about cheap indian restaurant \\

\midrule
E2E RG & In a task-oriented dialogue setting, generate a natural and helpful SYSTEM response to the USER query in the conversation provided in CONTEXT. You should follow the information provided in ACT to generate this answer. Do not mention that you are referring to a dataframe and don't overload the user with too many choices. You can perform \{TASK\}. \newline \newline
You can follow these examples: \newline
\{EXAMPLES\} \newline \newline
CONTEXT: \{CONTEXT\} \newline
ACT: \{ACT\} \newline
SYSTEM: 
 & In a task-oriented dialogue setting, generate a natural and helpful SYSTEM response to the USER query in the conversation provided in CONTEXT. You should follow the information provided in ACT to generate this answer. Do not mention that you are referring to a dataframe and don't overload the user with too many choices. You can perform restaurant booking. \newline \newline
You can follow these examples: \newline
USER: I need a place to fish. \newline
SYSTEM: Any preference in the type of fish? \newline
USER: Preferably salmons, but sardines are also fine. \newline
ACT: Blue Lake, 37th Avenue \newline
SYSTEM: How about in blue lake, 37th avenue? \newline \newline

CONTEXT: \newline
USER: Hi, any indian restaurants here? \newline
SYSTEM: Yes, we have Indian Express in the intermediate pricerange, would you want to know more? \newline
USER: Do you have any restaurant in the cheap pricerange? \newline
ACT: Royal Naan \newline
SYSTEM: 
 \\
\bottomrule
\end{tabular}
\caption{Templates and samples for the prompt used in the $RG_{naive}$, Proxy BS, initial KB interaction and E2E RG settings.}
\label{tab:prompt_templates}
\end{table*}

\end{document}